\pdfoutput=1
\documentclass[letterpaper,10pt,conference]{IEEEtran}

\usepackage[utf8x]{inputenc}
\usepackage[T1]{fontenc}
\usepackage[english]{babel}
\usepackage{amsmath}
\usepackage{amsfonts}
\usepackage{amssymb}
\usepackage{amsthm}
\usepackage{bm}
\usepackage{xspace}
\usepackage[boxruled]{algorithm2e}
\usepackage{mathrsfs}
\usepackage{multirow}
\usepackage{eurosym}
\usepackage{dsfont}
\usepackage[normalem]{ulem} 
\usepackage{balance}
\usepackage{soul}
\usepackage{pdfpages}
\usepackage{graphicx}
\usepackage{todonotes}
\usepackage{url}
\usepackage{siunitx}
\usepackage{stfloats}
\usepackage{array}
\usepackage{adjustbox}
\usepackage{float}
\usepackage{enumitem}
\usepackage[normalem]{ulem}
\setitemize{leftmargin=*}
\usepackage{balance}
\usepackage[numbers,sort&compress]{natbib}
\usepackage{etoolbox}
\apptocmd{\sloppy}{\hbadness 10000\relax}{}{}

\newcolumntype{L}[1]{>{\raggedright\let\newline\\\arraybackslash\hspace{0pt}}m{#1}}
\newcolumntype{C}[1]{>{\centering\let\newline\\\arraybackslash\hspace{0pt}}m{#1}}
\newcolumntype{R}[1]{>{\raggedleft\let\newline\\\arraybackslash\hspace{0pt}}m{#1}}

\newcommand{\T}{\mathsf{T}}
\newcommand{\norm}[1]{\left\lVert#1\right\rVert}

\newcommand{\mat}[1]{\begin{bmatrix}#1\end{bmatrix}}

\title{Feasibility Retargeting for Multi-contact Teleoperation and Physical Interaction}
\author{
\IEEEauthorblockN{Quentin Rouxel\IEEEauthorrefmark{1}, Ruoshi Wen\IEEEauthorrefmark{2}, Zhibin Li\IEEEauthorrefmark{3}, Carlo Tiseo\IEEEauthorrefmark{4}, Jean-Baptiste Mouret\IEEEauthorrefmark{1}, Serena Ivaldi\IEEEauthorrefmark{1}}
\IEEEauthorblockA{\IEEEauthorrefmark{1}Inria, CNRS, Universit\'e de Lorraine, France}
\IEEEauthorblockA{\IEEEauthorrefmark{2}Institute for Perception, Action, and Behaviour, School of Informatics, University of Edinburgh, UK}
\IEEEauthorblockA{\IEEEauthorrefmark{3}Department of Computer Science, University College London, UK}
\IEEEauthorblockA{\IEEEauthorrefmark{4}School of Engineering and Informatics, University of Sussex, UK}
}

\begin{document}
\thanks{
Quentin Rouxel, Jean-Baptiste Mouret, and Serena Ivaldi are with Inria, CNRS, Universit\'e de Lorraine, France.
Ruoshi Wen is with the Institute for Perception, Action, and Behaviour, School of Informatics, University of Edinburgh, UK.
Zhibin Li is with the Department of Computer Science, University College London, UK.
Carlo Tiseo is with the School of Engineering and Informatics, University of Sussex, UK.
}

\maketitle


\begin{abstract}
This short paper outlines two recent works \cite{seiko, seiko_franka} on multi-contact teleoperation and the development of the SEIKO (Sequential Equilibrium Inverse Kinematic Optimization) framework. SEIKO adapts commands from the operator in real-time and ensures that the reference configuration sent to the underlying controller is feasible. Additionally, an admittance scheme is used to implement physical interaction, which is then combined with the operator's command and retargeted. SEIKO has been applied in simulations on various robots, including humanoid and quadruped robots designed for loco-manipulation. Furthermore, SEIKO has been tested on real hardware for bimanual heavy object carrying tasks.
\end{abstract}



\section{Introduction}

As avatars become more capable and human-like, safety concerns arise due to the risk of falls and harm to the robot or its environment from operator errors \cite{johnson2015team}. The complexity of understanding the robot's balance and kinematics makes it necessary to automatically filter and retarget commands to ensure safe and feasible movements. Constraints include balance, joint position limits, stability conditions \cite{caron2015stability} for contact points (i.e. no pulling, no slipping, and no tilting), actuator torque limits. Our approach ensures that the desired configuration sent to the low-level controller always meets these feasibility constraints, enhancing safety and reducing the risk of harm.



Humans rely on contacts with their hands and feet for daily tasks. It enhances stability, increases the reaching distance for objects that are further away, and allows for a greater maximum pushing force when performing tasks such as opening doors. However, achieving balance and stability on uneven surfaces and with multiple contacts poses unique challenges \cite{khatib1, khatib2,di2016multi,shigematsu2019generating,bouyarmane2012humanoid,brossette2018multicontact}. The kinematic-only COM projection and ZMP criteria used for flat ground balance is insufficient \cite{abi2018humanoid,ishiguro2017bipedal}, as contact forces have infinite solutions, and the robot's force distribution can change with and without movement. To ensure stability, we consider the mutual influence of posture and contact force distribution while enforcing the appropriate constraints.


\begin{figure}[t]
    \centering
    \includegraphics[width=0.25\linewidth]{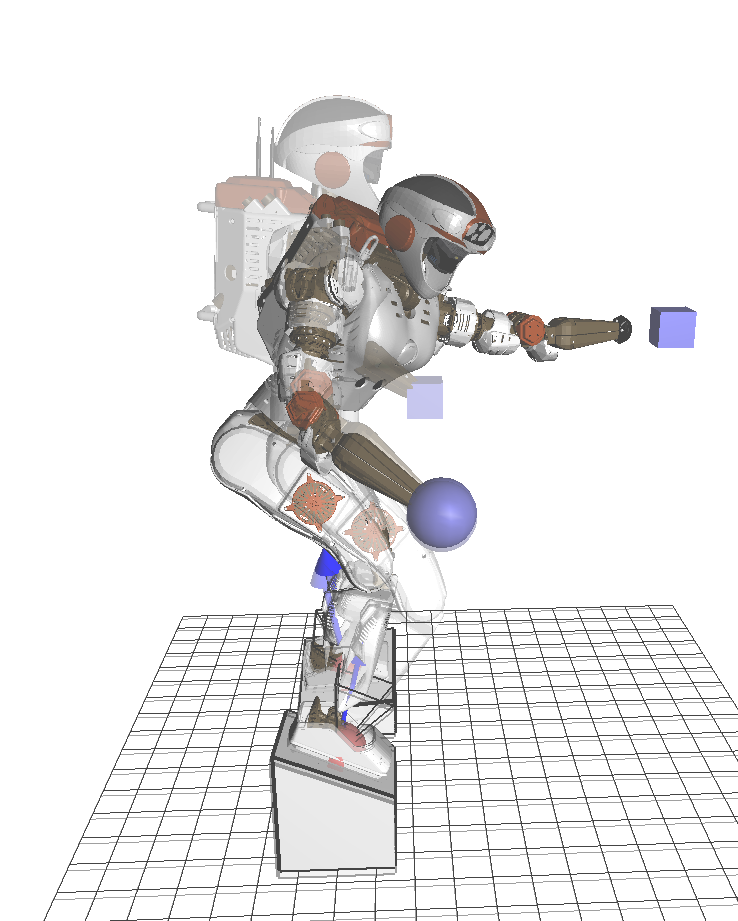}
    \includegraphics[width=0.25\linewidth]{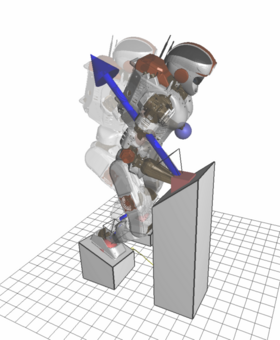}
    \includegraphics[trim=3.0cm 0.0cm 6.0cm 0.0cm,clip,width=0.25\linewidth]{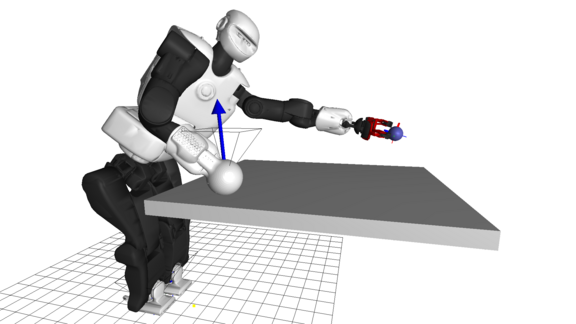}
    \includegraphics[trim=0.0cm 0.0cm 0.0cm 0.5cm,clip,width=0.35\linewidth]{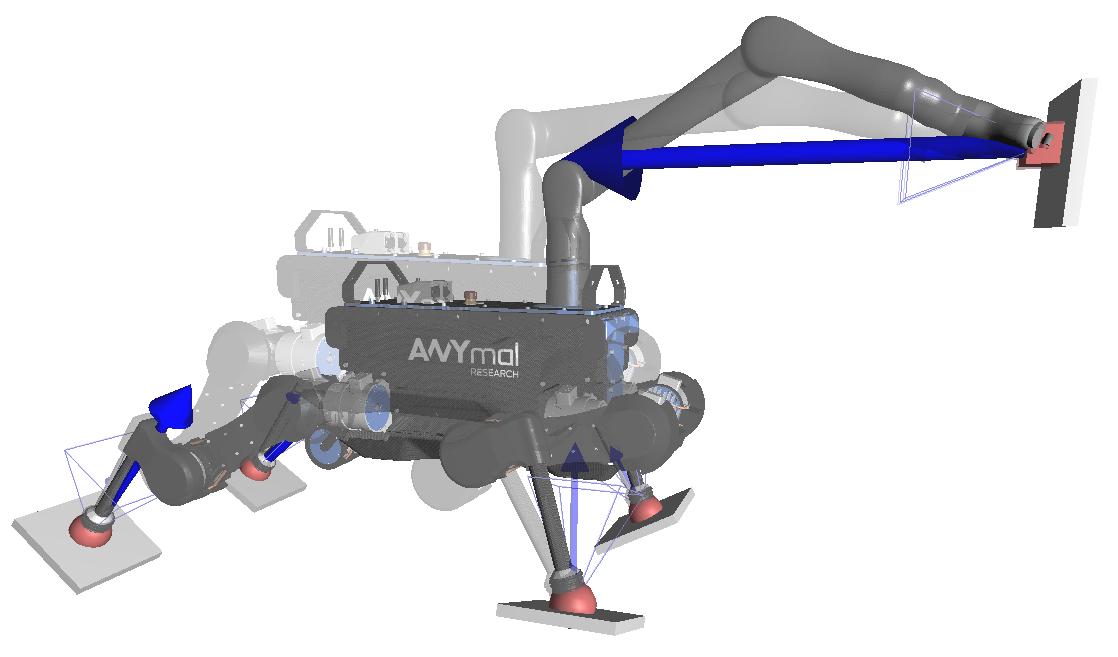}
    \includegraphics[trim=0.0cm 0.0cm 0.0cm 0.5cm,clip,width=0.35\linewidth]{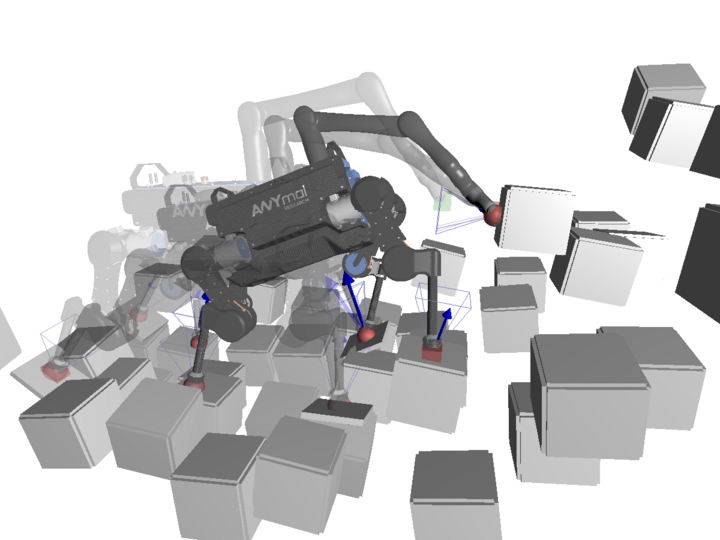}
    \includegraphics[trim=1cm 0 0.0cm 0.0cm,clip,width=0.35\linewidth]{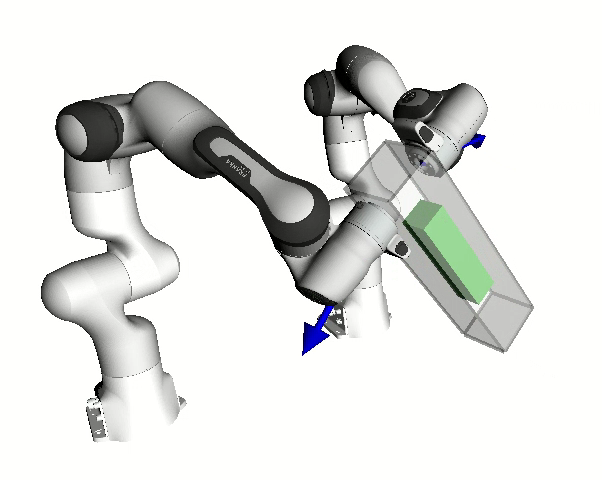}
    \includegraphics[trim=0cm 0 0.0cm 1.0cm,clip,width=0.35\linewidth]{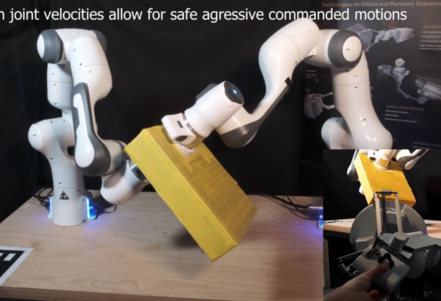}
    \caption{
        Multi-contact configurations computed by SEIKO during teleoperated tasks on Valkyrie and Talos humanoid robots, on ANYmal quadruped robot and on dual arms bimanual Franka robot. 
    }
    \label{fig:all_robots}
    \vspace{-5mm}
\end{figure}


Previous works that studied multi-contact and humanoid teleoperation \cite{survey} primarily focused on contact surfaces co-planar to the ground, whereas this work addresses the more general case of uneven surfaces and emphasizes smooth contact switching and feasibility boundary enforcement. While similar nonlinear optimizations were used for offline planning \cite{shigematsu2019generating} and posture generation \cite{brossette2018multicontact}, this work specifically prioritizes computational efficiency to enable real-time online reactive teleoperation.

\vspace{-2mm}
\section{Retargeting Optimization}

This short paper summarizes the two works \cite{seiko} and \cite{seiko_franka} that developed the SEIKO framework for Sequential Equilibrium Inverse Kinematic Optimization. SEIKO adapts operator commands to satisfy feasibility constraints and resolves posture and contact force redundancies for intuitive and robust avatar command. As depicted in Fig.~\ref{fig:all_robots}, the method has been applied in simulation and on real hardware on several robots.

SEIKO is a model-based method for multi-contact tasks on robots with multiple limbs and posture redundancy, including both fixed and floating base robots. It ensures the balance of either the robot itself through multiple supporting contacts or of an object being manipulated with multiple grasping points.

SEIKO assumes quasi-static conditions, making the problem instantaneous and avoiding consideration of future nonlinear dynamics. It optimizes nonlinear desired robot kinematics $\bm{q}^d$ and contact forces $\bm{\lambda}^d$ using a Sequential Quadratic Programming (SQP) scheme, solving a Quadratic Program (QP) at each time step:

\vspace{-5mm}
\begin{equation}
\begin{aligned}
    & \underset{\dot{\bm{x}}}{\text{min}}~\norm{\bm{C}_{\text{cost}}(\bm{x}_t) \dot{\bm{x}} -  \bm{c}_{\text{cost}}(\bm{x}_t)}^2_{\bm{w}} \text{~s.t.}\\
    & \bm{C}_{\text{eq}}(\bm{x}_t) \dot{\bm{x}} + \bm{c}_{\text{eq}}(\bm{x}_t) = \bm{0},~~ \bm{C}_{\text{ineq}}(\bm{x}_t) \dot{\bm{x}} + \bm{c}_{\text{ineq}}(\bm{x}_t) \geqslant \bm{0} \\
    & \text{where~} \bm{x}_t = \mat{\bm{q}^d \\ \bm{\lambda}^d},
    \dot{\bm{x}} = \mat{\dot{\bm{q}} \\ \dot{\bm{\lambda}} }, \\
\end{aligned}
\end{equation}

Here, $\bm{x}_t$ is the current desired configuration, and the incremental change $\dot{\bm{x}}$ is the decision variable of the QP. At each time step, the QP is solved and the desired configuration is integrated $\bm{x}_{t+1} = \bm{x}_t + \dot{\bm{x}}\Delta t$. The QP optimizes various objectives $\bm{C}_{\text{cost}},\bm{c}_{\text{cost}}$, including target effector poses, joint torques, and contact force minimization, while ensuring feasibility through equality and inequality constraints $\bm{C}_{\text{eq}},\bm{c}_{\text{eq}},\bm{C}_{\text{ineq}},\bm{c}_{\text{ineq}}$. Pinocchio library \cite{carpentier2019pinocchio} is used to compute the analytical partial derivatives of the static equation of motion to formulate $\bm{C}_{\text{eq}},\bm{c}_{\text{eq}}$: $\bm{G}(\bm{q}) = \bm{S} \bm{\tau} + \bm{J}(\bm{q})^\T \bm{\lambda}$ where $\bm{G}(\bm{q})$ is the gravity vector, $\bm{\tau}$ is the joint torque, $\bm{J}(\bm{q})$ is the stacked Jacobian matrix of enabled contacts, and $\bm{S}$ is the selection matrix accounting for the floating base. The optimization runs in real-time (one iteration is under $1$ms), converges quickly and continuously retargets operator commands to feasible desired configurations of posture and contact forces.

SEIKO generates anthropomorphic postures for humanoid avatars by minimizing joint torques. \cite{saeid1, saeid2} optimize directly the position of the COM to maximize robustness margins against disturbances during multi-contact sliding tasks. In contrast, our work solves for velocities, focuses on achieving stability close to the feasibility limits in order to fully exploit the capabilities of the system and proposes a smooth contact switching procedure that is critical for complex loco-manipulation tasks. While SEIKO has successfully accomplished teleoperated tasks at slow or moderate speeds, it is not suitable for fast motions that consider inertial effects. Contact locations are not optimized, and the operator must select the position and sequence of contact points.

\section{Interface, Interaction and Architecture}

\begin{figure}[t!]
    \centering
    \includegraphics[width=0.59\linewidth]{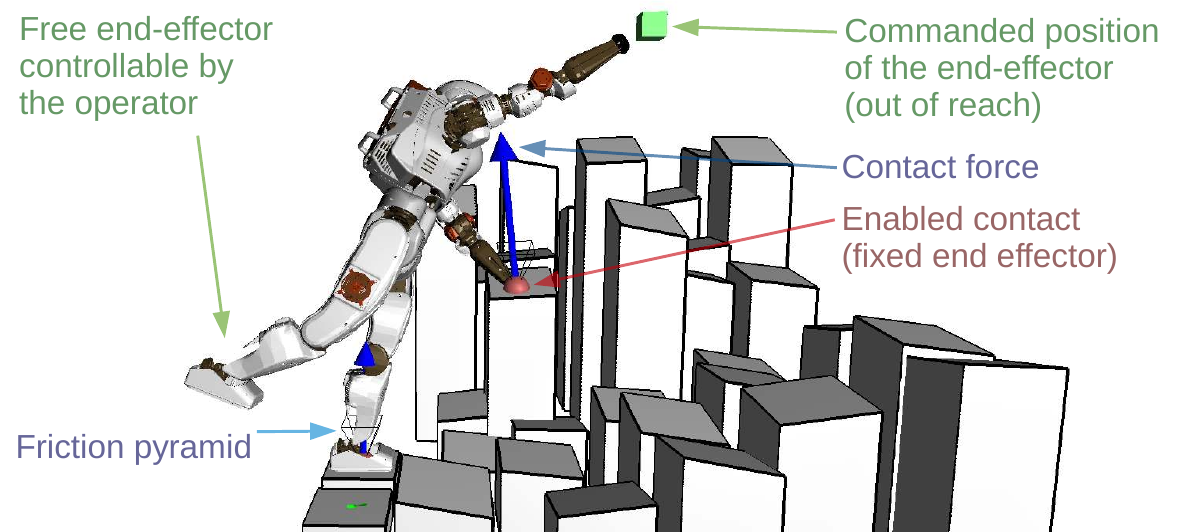}
    \includegraphics[trim=2cm 0 5cm 2.0cm,clip,width=0.29\linewidth]{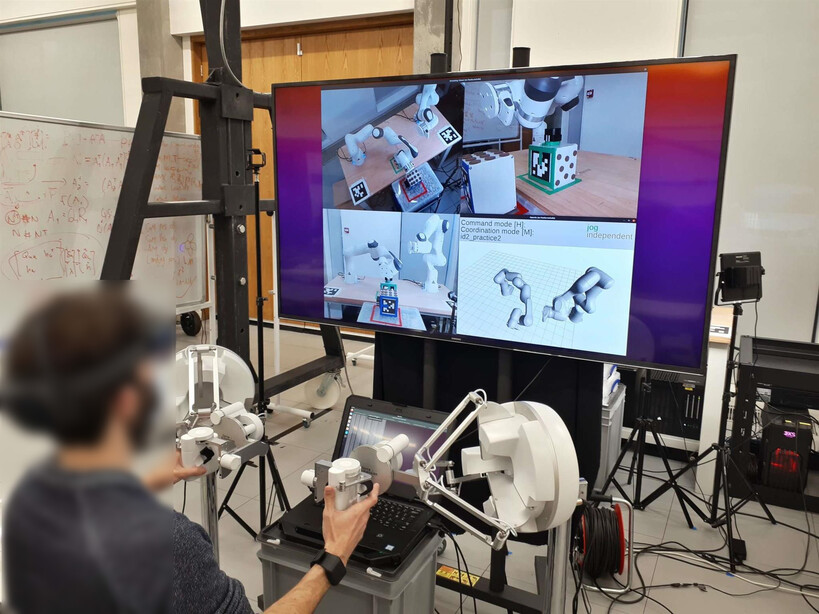}
    \caption{
        User interface for simulated multi-contact teleoperation (\textit{left}) where the operator commands the pose of free end-effectors and triggers contact switch. Teleoperation interface for remote bimanual docking and assembly task (\textit{right}).
    }
    \label{fig:interfaces}
    \vspace{-6mm}
\end{figure}

Designing human-robot interfaces for teleoperated multi-contact tasks is a complex challenge. Tracking the entire body of the operator as a reference \cite{MontecilloPuente2010OnRW,koenemann2014real,penco2017robust,darvish2019whole} is inconvenient for humanoid robots due to factors like operator fatigue, kinematic and inertia mismatch, and the need to replicate environmental contacts. Our approach, illustrated in Fig.~\ref{fig:interfaces}, commands the Cartesian pose of end-effectors and uses buttons to trigger contact switches.

\begin{figure}[t!]
    \centering
    \includegraphics[trim=0cm 2cm 0cm 0.0cm,clip,width=0.9\linewidth]{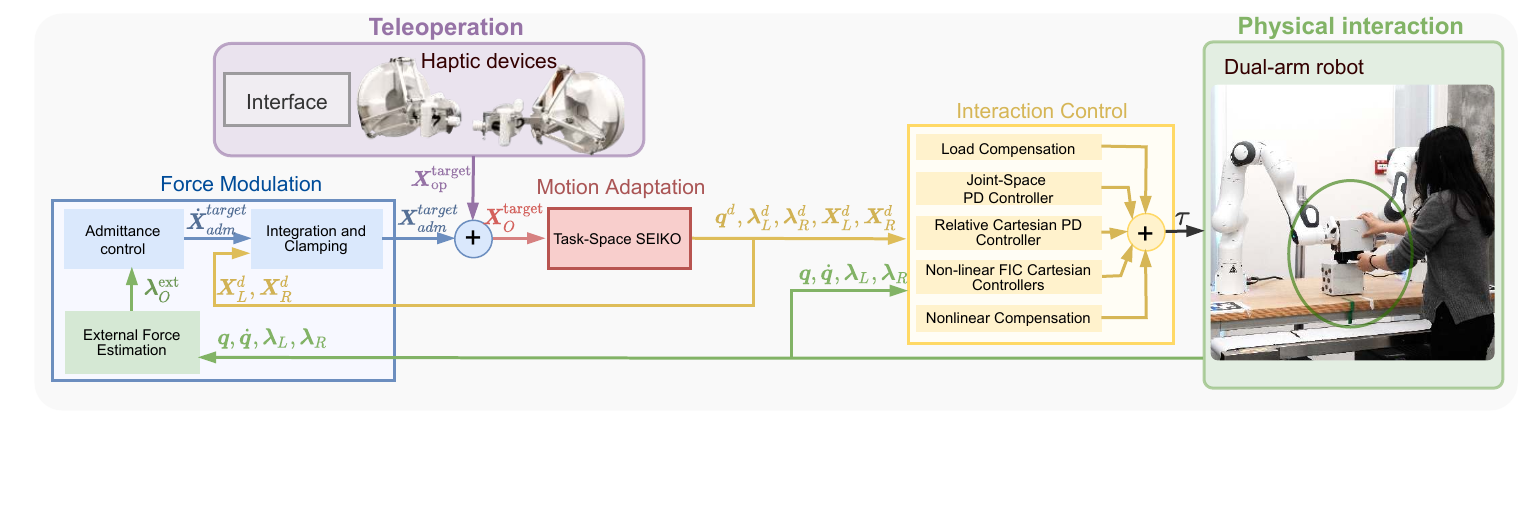}
    \caption{
        An admittance controller processes external forces into velocity commands, which are integrated into a relative reference pose. SEIKO adapts the combined teleoperation and physical interaction commands to satisfy system and task constraints, producing feasible desired posture and contact wrenches. The interaction controller realizes the desired configuration using passive impedance and load compensation controllers.
    }
    \label{fig:architecture}
    \vspace{-4mm}
\end{figure}

\begin{figure}[t!]
    \centering
    \includegraphics[width=0.9\linewidth]{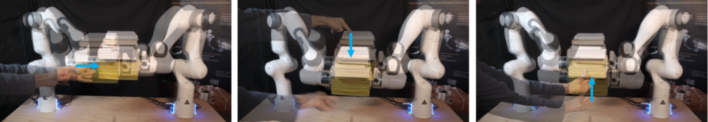}
    \caption{
        A stack of books is held and moved through physical interaction by a dual-arm robot and a local operator collaborating together.
    }
    \label{fig:physical_interraction}
    \vspace{-6mm}
\end{figure}

To perform collaborative tasks such as carrying a large object or being physically guided, robotic avatars should be able to interact with humans or their environment. Compliance at end-effectors is helpful for establishing contacts without creating disturbances. Our work achieves end-effector compliance using a velocity admittance scheme computed from external forces (Fig.~\ref{fig:architecture}), enabling a human collaborator to push and move the robot's effectors (Fig.\ref{fig:physical_interraction}). However, the challenge is to maintain feasibility constraints and prevent the robot from being pushed towards an infeasible or dangerous posture by resisting physical interaction when necessary. Our work demonstrates the implementation of safe physical interaction while superimposing the operator's pose commands.

A filtering pipeline including low-pass, maximum velocity, and acceleration filters, processes effector pose commands before SEIKO to further enhance robustness against aggressive motions and network communication jitter.

\section{Results and Conclusion}

The SEIKO framework has undergone extensive validation on humanoid, quadruped, and fixed-base dual-arm robots, with both plane and point contacts. We have utilized Gazebo and PyBullet simulators \cite{seiko} to test multi-contact tasks, such as extreme reaching beyond feasibility boundaries, pushing, contact switching, and traversal of complex and uneven terrain. We presented quantitative analysis showing that despite the quasi-static assumption, practical tasks with moderate speed were achievable. On humanoid Valkyrie, simulated hand velocity can reach $\SI{30}{cm/s}$ by trading off maximum reachable distance for more conservative postures. The framework has also been tested on real robots \cite{seiko_franka} for bimanual grasping and maneuvering of heavy objects in arbitrary ways, guiding the robot through physical interactions with stacks of unknown objects, collaborative part assembly, and insertion of industrial connectors. These results illustrated in our videos\footnote{\url{https://doi.org/10.1109/TMECH.2022.3152844/mm2}}\footnote{\url{https://youtu.be/A0bjCIIHyjQ}} highlight the versatility and robustness of the SEIKO framework for a wide range of robotic applications.

\section*{Acknowledgment}

This research is supported by the EPSRC Future AI and Robotics for Space (EP/R026092), ORCA (EP/R026173), NCNR (EP/R02572X), EU Horizon2020 project THING (ICT-2017-1), EU Horizon2020 project Harmony (101017008), and EU Horizon project euROBIN (101070596).


\bibliographystyle{IEEEtran}
\bibliography{references}

\end{document}